\documentclass[9pt,journal]{IEEEtran}

\usepackage{amsmath}
\usepackage{amssymb}
\usepackage{bm}
\usepackage{url}
\usepackage{multirow}
\usepackage{booktabs}
\usepackage[figuresright]{rotating}
\usepackage{oubraces}
\usepackage{mathtools}
\usepackage{subfigure}
\usepackage{cite}
\newcommand{\vc}[1]{\boldsymbol{\mathbf{#1}}}
\DeclareMathSymbol{\R}{\mathalpha}{AMSb}{"52}
\usepackage{color}
\definecolor{bostonuniversityred}{rgb}{0.8, 0.0, 0.0}
\definecolor{bostonuniversityred}{rgb}{0.0, 0.0, 0.0}
\begin{document}

\title{Adaptive Propagation Graph Convolutional Network}
\author{Indro Spinelli,~\IEEEmembership{Graduate Student Member,~IEEE}, Simone Scardapane, and Aurelio Uncini,~\IEEEmembership{Member,~IEEE}%
\thanks{Emails: \{firstname.lastname\}@uniroma1.it}
\thanks{Authors are with the Department of Information Engineering, Electronics and Telecommunications (DIET), Sapienza University of Rome, Italy}
}

\markboth{Paper published in \textbf{IEEE Transaction on Neural Networks and Learning Systems} (editorial preprint)}%
{Spinelli \MakeLowercase{\textit{et al.}}: Adaptive propagation GNN}

\maketitle
\begin{abstract}
Graph convolutional networks (GCNs) are a family of neural network models that perform inference on graph data by interleaving vertex-wise operations and message-passing exchanges across nodes. Concerning the latter, two key questions arise: (i) how to design a differentiable exchange protocol (e.g., a $1$-hop Laplacian smoothing in the original GCN), and (ii) how to characterize the trade-off in complexity with respect to the local updates. In this paper, we show that state-of-the-art results can be achieved by adapting the number of communication steps independently at every node. In particular, we endow each node with a halting unit (inspired by Graves' adaptive computation time \cite{graves2016adaptive}) that after every exchange decides whether to continue communicating or not. We show that the proposed adaptive propagation GCN (AP-GCN) achieves superior or similar results to the best proposed models so far on a number of benchmarks, while requiring a small overhead in terms of additional parameters. We also investigate a regularization term to enforce an explicit trade-off between communication and accuracy. The code for the AP-GCN experiments is released as an open-source library.
\end{abstract}

\begin{IEEEkeywords}
Graph neural network, Graph data, Convolutional network, Node classification
\end{IEEEkeywords}

\section{Introduction}
\label{sec:introduction}
\IEEEPARstart{D}{eep} learning has achieved remarkable success on a number of high-dimensional inputs, by properly designing architectural biases that can exploit their properties. This includes images (through convolutional filters) \cite{lecun2015deep}, {\color{bostonuniversityred}text \cite{yao2019graph}}, biomedical sequences \cite{webb2018deep}, and videos \cite{aytar2016soundnet}. A major research question, then, is how to replicate this success on other types of data, through the implementation of novel differentiable blocks adequate to them. Among the possibilities, \textit{graphs} represent one of the largest sources of data in the world, ranging from recommender systems \cite{berg2017graph} to biomedical applications \cite{gilmer2017neural}, social networks \cite{newman2002random}, computer programs \cite{allamanis2017learning}, knowledge bases \cite{schlichtkrull2017modeling}, and many others.

In its most general form, a graph is composed by a set of vertices connected by a series of edges representing, e.g., social connections, citations, or any form of relation. Graph neural networks (GNNs) \cite{bronstein2017geometric,wu2019comprehensive,bacciu2019gentle}, then, can be designed by interleaving local operations (defined on either individual nodes or edges) with communication steps, exploiting the graph topology to combine the local outputs. These architectures can then be exploited for a variety of tasks, ranging from node classification to edge prediction and path computation.

Among the different families of GNN models proposed over the last years, graph convolutional networks (GCN) \cite{kipf2017semi} have become a sort of \textit{de facto} standard for node and graph classification, representing one of the simplest (yet efficient) building blocks in the context of graph processing. GCN are built by interleaving vertex-wise operations, implemented via a single fully-connected layer, with a communication step exploiting the so-called Laplacian matrix of the graph. In practice, a single GCN layer provides a weighted combination of information across neighbors, representing a localized $1$-hop exchange of information.\footnote{In the paper, we use \textit{node} and \textit{vertex} as synonyms, and the same for \textit{communication} and \textit{propagation}.}

Taking the GCN layer as a fundamental building block, several research questions have received vast attention lately, most notably: (i) how to design more effective communication protocols, able to improve the accuracy of the GCN and potentially better leverage the structure of the graph \cite{levie2018cayleynets,bianchi2019graph,klicpera2019diffusion}; and (ii) how to trade-off the amount of local (vertex-wise) operations with the communication steps \cite{klicpera2018predict}. While we defer a complete overview of related works to Section \ref{sec:related_works}, we briefly mention two key results here. Firstly, \cite{li2018deeper} showed that the use of the Laplacian (a smoothing operator) has as consequence that repeated application of standard GCN layers tend to over-smooth the data, disallowing the possibility of naively stacking GCN layer to obtain extremely deep networks. Secondly, \cite{klicpera2018predict} showed that state-of-the-art results can be obtained by replacing the Laplacian communication step with a PageRank variation, as long as completely separating communication between nodes from the vertex-wise operations. We exploit both of these key results later on.

\subsection{Contributions of the paper}
We note that the vast majority of proposals to improve point (i) mentioned before consists in selecting a certain maximum number of communication steps $T$, and iterating a simple protocol for $T$ steps in order to diffuse the information across $T$-hop neighbors. In this paper, we ask the following research question: can the performance of GCN layers be improved, if the number of communication steps is allowed to vary \textit{independently} for \textit{each} vertex?

To answer this question, we propose a variation of GCN that we call adaptive propagation GCN (AP-GCN). In the AP-GCN (see Fig. \ref{fig:framework}) every vertex is endowed with an additional unit that outputs a value controlling whether communication should continue for another step (hence combining the information from neighbors farther away), or should stop, and the final value be kept for further processing. In order to implement this adaptive unit, we leverage previous work on adaptive computation time in recurrent neural networks \cite{graves2016adaptive} to design a differentiable method to learn this propagation strategy. On an extensive set of comparisons and benchmarks, we show that AP-GCN can reach state-of-the-art results, while the number of communication steps can vary significantly not only across datasets but also across individual vertexes. This is achieved with an extremely small overhead in terms of computational time and additional trainable parameters. In addition, we perform an large hyper-parameter analysis, showing that our method can provide a simple way to balance accuracy of the GCN with the number of propagation steps.

\subsection{Outline of the paper}

The rest of the paper is structured as follows. In Section \ref{sec:related_works} we describe more in-depth related works from the field of GCNs and GNNs, focusing in particular on several proposals describing how to design more complex propagation steps. Then, in Section \ref{sec:graph_cns} we introduce the GCN model and the way a deep network can be composed and trained from GCN blocks. Our proposed AP-GCN is first introduced in Section \ref{sec:proposed} and then tested in Section \ref{sec:experimental_results}. We conclude with some general remarks in Section \ref{sec:conclusion}.

\section{Related works}
\label{sec:related_works}
GCNs belong to the class of spectral graph neural networks, which are based on graph signal processing (GSP) tools \cite{sandryhaila2014big,ortega2018graph,romero2017kernel,di2016adaptive}. GSP allows to define a Fourier transform over graphs by exploiting the eigen-decomposition of the so-called graph Laplacian. The first application of this theory to graph NNs was in \cite{bruna2014spectral}. This approach, however, was both computationally heavy and not spatially localized, meaning that each node-wise update depended on the entire graph structure. Later proposals \cite{defferrard2016convolutional} showed that by properly restricting the class of filters applied in the frequency domain, one could obtain simpler formulation that were also spatially localized in the graph domain. Polynomial filters \cite{defferrard2016convolutional} can be implemented via $T$-hop exchanges on the graph, but they require to select \textit{a priori} a valid $T$ for all the vertices. The GCN, introduced in \cite{kipf2017semi}, showed that state-of-the-art results could be obtained even with simpler linear (i.e., $1$-hop) operations. However, they failed to build deeper architectures (i.e., $>2$ GCN layers) in practice.

{\color{bostonuniversityred}The authors of} \cite{li2018deeper}, formally analyzed the properties of the GCN, showing that the difficulty of building deeper networks could depend from the over-smoothing of the data due to a repeated application of the Laplacian operator. Further analyses and the need to consider higher-order structures in GNNs were provided by \cite{morris2019weisfeiler}, showing that GCNs are equivalent to the so-called 1-dimensional Weisfeiler-Leman graph isomorphism heuristic. Several recent papers have proposed to avoid some of these shortcomings by using different types of propagation methods, most notably PageRank variations \cite{klicpera2018predict,klicpera2019diffusion}. 

In this paper we explore an orthogonal idea, where we hypothesize that performance can be improved not only by modifying the existing propagation method, but by allowing each node to vary the amount of communication independently from the others, in an adaptive fashion. Jumping knowledge (JK) networks \cite{xu2018representation} and GeniePath \cite{liu2019geniepath} achieve something similar by exploiting an additional network aggregation component (e.g., an LSTM network) after multiple diffusion steps, however, they fail to reach state-of-the-art results \cite{klicpera2019diffusion}.

Finally, we underline that we focus on GCN in this paper, but alternative models for graph neural networks have been devised, including those from \cite{scarselli2008graph}, graph attention networks \cite{velivckovic2017graph}, graph embeddings, and others. We refer to multiple recent surveys on the topic for more information \cite{wu2019comprehensive,bacciu2019gentle}.

\section{Graph convolutional neural networks}
\label{sec:graph_cns}

\subsection{Graph definitions}
\label{subsec:graph_definitions}

Consider a generic undirected graph $\mathcal{G} = (\mathcal{V}, \mathcal{E})$, where $\mathcal{V} = \left\{1, \ldots, n\right\}$ is the set of node indexes, and $\mathcal{V} = \left\{(i, j) \; \vert \; i, j \in \mathcal{V}\right\}$ is the set of arcs (\textit{edges}) connecting pairs of nodes. The meaning of a single node or edge depends on the application. For example, a classic setup in text classification encodes each text as a node \cite{kipf2017semi}, and a citation among two texts as an arc in the corresponding graph.

Connectivity in the graph can be summarized in the adjacency matrix $\vc{A} \in \left\{0, 1\right\}^{n\times n}$. From this, we can define the diagonal degree matrix $\vc{D}$ where $D_{ii} = \sum_j A_{ij}$, and the Laplacian matrix $\vc{L} = \vc{D} - \vc{A}$. In the context of GNNs, the Laplacian is generally used in its \textit{normalized} form $\widehat{\vc{L}} = \vc{D}^{-1/2}\vc{L}\vc{D}^{-1/2}$. As we will see, the Laplacian operators can be used to define (normalized) 1-hop communication protocols across the graph.

In the context of inference over graphs, we suppose that node $i$ is endowed with a vector $\vc{x}_i \in \mathbb{R}^d$ of features. For tasks of node classification \cite{kipf2017semi}, we also know a desired label $y_i$ for a subset $\mathcal{T} \subset \mathcal{V}$ of nodes, and we wish to infer the labels for the remaining nodes. Graph classification is easily handled by considering sets of graphs defined as above, with a single label associated to every graph, e.g., \cite{gilmer2017neural}. While we focus on node / graph classification in the rest of the paper, the techniques we introduce in the next section can further be extended by considering edge features $\vc{v}_{ij}$, global graph features \cite{battaglia2018relational}, and applied to other tasks such as edge classification \cite{berg2017graph}. We {\color{bostonuniversityred}will} return on this {\color{bostonuniversityred}argument} in Section \ref{subsec:design_train_deep_gcns}.

\subsection{Graph convolutional networks}
\label{subsec:gcn}
The basic idea of GCNs is to combine local (node-wise) updates with suitable message passing across the graph, following the graph topology. In particular, consider the $n \times d$ matrix $\vc{X}$ collecting all node features for the entire graph. A generic GCN layer can be written as \cite{kipf2017semi}:
\begin{equation}
    \vc{H} = \phi \left( \widehat{\vc{L}} \vc{X} \vc{W} + \vc{b} \right) \,,
    \label{eq:gcn_layer}
\end{equation}
where $\phi$ is an element-wise nonlinearity (such as the ReLU $\phi(\cdot) = \max\left(0, \cdot\right)$), $\widehat{\vc{L}}$ is the normalized Laplacian defined above, and $\vc{W}$ and $\vc{b}$ are the learnable parameters of the layer. More in general, the Laplacian matrix can be renormalized in different ways (see \cite{kipf2017semi}) or substituted with any appropriate shift operator defined on the graph.

The name GCN derives from an interpretation of {\color{bostonuniversityred}Equation} \eqref{eq:gcn_layer} in terms of GSP \cite{sandryhaila2014big}, as described in Section \ref{sec:related_works}. A graph Fourier transform can be defined for the graph by considering the eigen-decomposition of the Laplacian matrix \cite{di2016adaptive}. In this context, {\color{bostonuniversityred}Equation} \eqref{eq:gcn_layer} can be shown to be equivalent to a graph convolution implemented with a linear filter \cite{kipf2017semi}. Because its implementation requires only 1-hop exhanges across neighbours, the GCN is also an example of a message-passing neural network (MPNN) \cite{bacciu2019gentle}. 

These two interpretations bring forth two classes of extensions for the basic model in {\color{bostonuniversityred}Equation} \eqref{eq:gcn_layer}, which we comment on to the extent that they relate to our proposed method. Firstly, under a GSP interpretation, it makes sense to substitute the linear filtering operation with a more complex filter.{\color{bostonuniversityred}\footnote{In fact, as we described in Section \ref{sec:related_works}, some of these works predate the introduction of the GCN itself.} In particular, polynomial filters can be implemented by combining information from higher-order neighborhoods of each node, depending on the degree of the polynomial \cite{gama2020graphs}.} For example, Chebyshev filters \cite{defferrard2016convolutional} result in the following layer (omitting biases for simplicity):
\begin{equation}
    \vc{H} = \phi \left( \sum_{k=1}^K T_k(\widehat{\vc{L}})\vc{X}\vc{W}_k \right) \,,
    \label{eq:chebyshev_layer}
\end{equation}
where $T_k(s)$ is defined recursively as $T_k(s) = 2sT_{k-1}(s) - T_{k-2}(s)$, and the layer has a number of adaptable matrices $\left\{\vc{W}_k\right\}_{k=1}^K$ that depend on the user-defined hyper-parameter $K$. Setting $K$ corresponds to selecting a `depth' for the information being propagated. For example, setting $K=2$ propagates information across 2-hop neighbors, while $K=1$ {\color{bostonuniversityred}is (almost) equivalent} to the GCN described above. This decision, however, must be made beforehand by the user, or the parameter must be fine-tuned accordingly.


\begin{figure*}
    \centering
     \includegraphics[width=1.9\columnwidth]{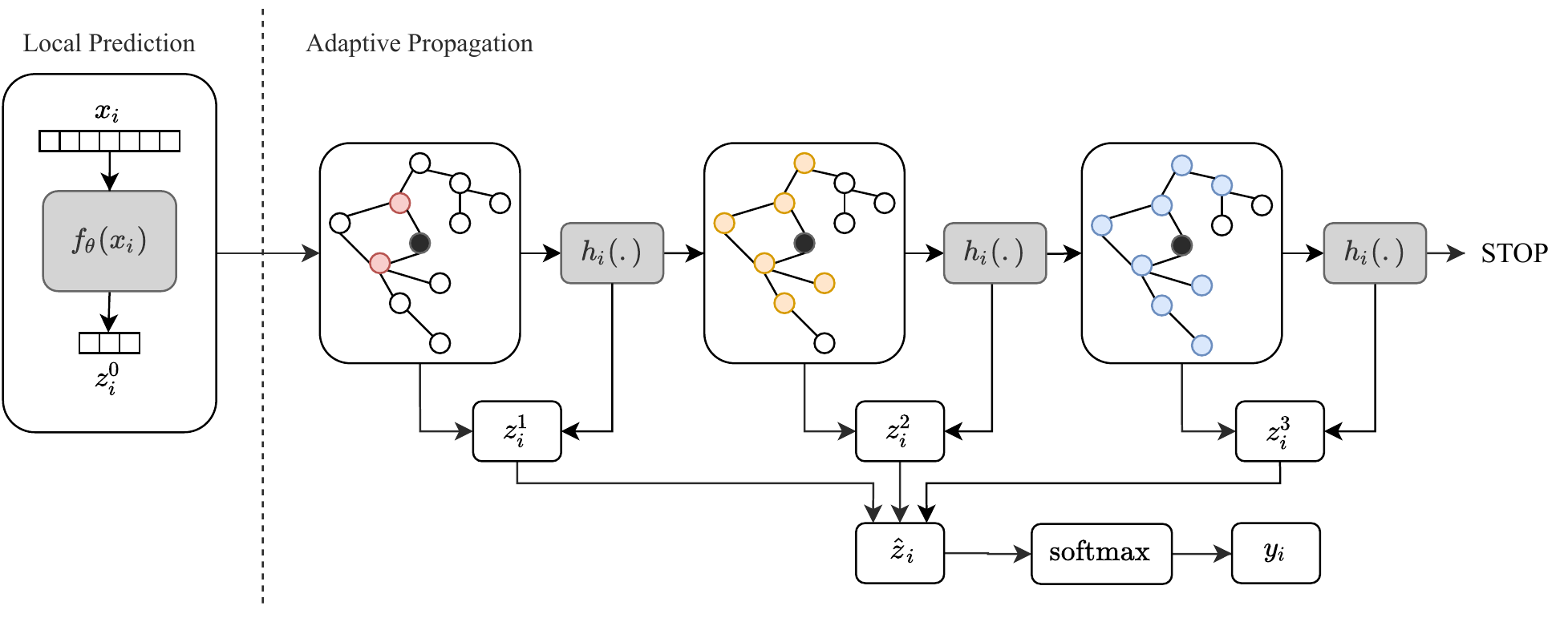}
    \caption{Schematics of the proposed framework.}
    \label{fig:framework}
\end{figure*}

Under the more general interpretation of {\color{bostonuniversityred}Equation} \eqref{eq:gcn_layer} as a MPNN, however, we are not restricted to considering filtering operations. In fact, the most general extension of {\color{bostonuniversityred}Equation} \eqref{eq:gcn_layer} becomes  (expressed for simplicity for a single node $i$) \cite{bacciu2019gentle}:
\begin{equation}
    \vc{h}_i = \Psi\left( \left\{ \psi(\vc{x}_j) \;\vert\; j \in \mathcal{N}_i \right\} \right) \,,
    \label{eq:mpnn}
\end{equation}
where $\Psi$ is a permutation-invariant function, $\psi$ a node-wise update, and $\mathcal{N}_i$ is the {\color{bostonuniversityred} neighborhood of node $i$ (where in general $i \in \mathcal{N}_i$)}. Selecting $\psi(\vc{x}) = \vc{W}^T\vc{x}$ and $\Psi(\left\{\psi(\vc{x}_i)\right\}) = \sum_{j}\widehat{L}_{ij}\psi(\vc{x}_j)$ recovers the previous GCN formulation. More in general, both $\Psi$ and $\psi$ can be implemented as generic neural networks or any other differentiable mechanism. Most notably, \cite{klicpera2018predict} proposes the use of (approximate) PageRank protocols for the propagation step to counteract the oversmoothing effect of repeated applications of the Laplacian matrix \cite{li2018deeper}, although the maximum number of propagation steps must still be selected \textit{a priori} by the user.

{\color{bostonuniversityred}Interestingly, PageRank propagation \cite{klicpera2018predict} and the closely-related ARMA models \cite{bianchi2019graph}, can be understood as approximating \textit{rational} filters on the graph \cite{chen2020bridging}, that are in general more expressive than linear or polynomial filters.}

\subsection{Designing and training deep GCNs}
\label{subsec:design_train_deep_gcns}
In the spirit of classical deep networks, the basic building blocks described in the previous section can be composed to design deeper architectures. For example, a network for binary classification with a single hidden layer and one output layer, both implemented according to {\color{bostonuniversityred}Equation} \eqref{eq:gcn_layer}, is defined by:
\begin{equation}
    \vc{y} = \sigma\left( \widehat{\vc{L}} \cdot \phi\left( \widehat{\vc{L}}\vc{X}\vc{W} + \vc{b} \right) \vc{v} + c \right) \,,
\end{equation}
where the adaptable weights are $\vc{W}$, $\vc{v}$, $\vc{b}$ and $c$. A more recent line of reasoning, popularized by \cite{klicpera2019diffusion}, is to implement architectures in the form {\color{bostonuniversityred}Equation} \eqref{eq:mpnn}, making both $\psi$, $\Psi$ deeper networks, but without interleaving multiple node-wise and propagation steps. We follow this design principle here, as we have found it to perform better empirically.

Once a specific network $f$ has been designed, its optimization follows the same strategies as for other deep networks. For example, for node classification (as described in Section \ref{subsec:graph_definitions}){\color{bostonuniversityred},} we optimize the network with a cross-entropy loss on the known node labels:
\begin{equation}
    f^* = \arg\min \left\{ \sum_{i \in \mathcal{T}} y_i \cdot \log\left(f(\vc{x}_i)\right) \right\} \,.
    \label{eq:node_classification_opt}
\end{equation}
Note, however, that differently from standard neural networks, the output of $f(\vc{x}_i)$ will depend on several other nodes, depending on the specific architecture. For this reason, {\color{bostonuniversityred}Equation} \eqref{eq:node_classification_opt} is harder to solve efficiently in a stochastic fashion \cite{sato2019constant}.

\section{Proposed adaptive propagation protocol}
\label{sec:proposed}

In the previous sections, we analyzed the motivation for having graph modules with complex diffusion steps across the graph. However, the vast majority of proposals has considered a single, maximum number of communication steps that is shared for all the nodes in the graph (e.g., the number $K$ in {\color{bostonuniversityred}Equation} \eqref{eq:chebyshev_layer}). In this section we introduce a novel variation of GCN wherein (i) the number of communication steps is selected independently for every node, and (ii) this number is adapted and computed on-the-fly during training. To the best of our knowledge, our proposed Adaptive Propagation GCN (AP-GCN) is the only model in the literature combining these two properties.

Our AP-GCN framework is summarized in Fig. \ref{fig:framework}. Considering the notation in {\color{bostonuniversityred}Equation} \eqref{eq:mpnn}, we separate the node-wise operations $\psi$ from the propagation step $\Psi$. The former is implemented with a generic NN applied on a single node $\mathbf{z}_j = \psi(\mathbf{x}_j)$, described on the left part of Fig. \ref{fig:framework}. This embedding is then used as the starting seed for a propagation step $\Psi$ which is done iteratively:
\begin{align}
    \mathbf{z}_i^0 & = \mathbf{z}_i \nonumber \\
    \mathbf{z}_i^1 & = \text{propagate}(\left\{\mathbf{z}_j^0 \;\;\vert\;\; j \in \mathcal{N}_i\right\}) \nonumber \\
    \mathbf{z}_i^2 & = \text{propagate}(\left\{\mathbf{z}_j^1 \;\;\vert\;\; j \in \mathcal{N}_i\right\}) \nonumber \\
    & \ldots \nonumber
\end{align}
Key to our proposal, the number of propagation steps depends on the index of node $i$ and it is computed adaptively while propagating. The mechanism to implement this is inspired by the adaptive computation time in RNNs \cite{graves2016adaptive}.

First, we endow each node with a linear binary classifier acting as a `halting unit' for the propagation process. After the generic iteration $k$ of propagation, we compute node-wise:
\begin{equation}
    h_i^k = \sigma\left( \mathbf{Q}\mathbf{z}_i^k + q \right) \,,
\end{equation}
where $\mathbf{Q}$ and $q$ are trainable parameters. The value $h_i^k$ describes the probability that the node should stop after the current iteration. In order to ensure that the number of propagation steps remains reasonable, following \cite{graves2016adaptive} we adopt two techniques. Firstly, we fix a maximum number of iterations $T$. Secondly, we use the running sum of the halting values to define a budget for the propagation process:
\begin{equation}
    K_i = \min \left\{k^{\prime}: \sum_{k=1}^{k^{\prime}} h_i^k > =1-\epsilon\right\} \,,
\end{equation}
where $\epsilon$ is a hyper-parameter, generally set to a small value, that ensures that the process can terminate also after a single update. Whenever $k = K_i$, the budget is reached and the propagation stops for node $i$ at iteration $k$. {\color{bostonuniversityred} We combine the halting probabilities as follows:}
%
\begin{equation}
    p_i^k = \begin{cases} R_i = 1 - \sum_{k=1}^{K_i-1} h_i^k\:  & \text { if } \:  k=K_i \text{ or } k=T \\ \sum_{k=1}^{K_i} h_i^k \: & \text{ otherwise} \end{cases} \,,
\end{equation}
%

%
{\color{bostonuniversityred} In this way the sequence $\left\{p_i\right\}$ forms a valid cumulative distribution for the halting probabilities $\left\{h_i\right\}$.}
By exploiting it, instead of using the latest value in the propagation, we can adaptively combine the information at every step for free:
%
\begin{equation}
\label{eq:z}
    \widehat{\mathbf{z}}_i = \frac{1}{K_i}\sum_{k=1}^{K_i} p_i^k \mathbf{z}_i^k + (1-p_i^k) \mathbf{z}_i^{k-1} \,.
\end{equation}
$\widehat{\mathbf{z}}_i$ is now the final output for node $i$.

The number of propagation steps can be controlled by the definition of a propagation cost {\color{bostonuniversityred}  $\mathcal{S}_i$, similarly to \cite{graves2016adaptive}, which represents the amount of propagation steps needed for the update of the $i$-th node:}
\begin{equation}
    \mathcal{S}_i = K_i + R_i \,.
\end{equation}
Denoting by $\mathcal{L}$ the loss term in {\color{bostonuniversityred}Equation} \eqref{eq:node_classification_opt}, this term is added to be minimized, weighed by a propagation penalty $\alpha$:
\begin{equation}
    \widehat{\mathcal{L}} =\mathcal{L} + \alpha \sum_{i \in \mathcal{V}} \mathcal{S}_i \,.
\end{equation}
The propagation penalty is responsible for the trade-off between computation time and accuracy. Moreover, it regulates how `easily' the information spreads on the graph.
In practice, the optimization of the halting unit is performed in an alternate fashion once every $L$ steps of the main network (in our experiments, $L=5$).

\section{Experimental results}
\label{sec:experimental_results}

\subsection{Experimental setup}

We used the same experimental setup proposed in \cite{klicpera2018predict} which aims to reduce experimental bias. This setup has shown that many advantages reported by recent works vanish under this statistically rigorous evaluation. The first step in this process is the subdivision in visible and invisible sets. The invisible set will serve as a test set and will be used only once to report the final performance. The visible set is subdivided in a training set with $N$ nodes per class and an early stopping set for model selection. A validation set containing the remaining nodes of the visible set is used for hyper-parameters tuning. These splits are determined using the same $20$ seeds used in \cite{klicpera2018predict} and each experiment is run with $5$ different initialization of the weights leading to a total of $100$ experiments per dataset.

{\color{bostonuniversityred} We perform a first evaluation over three citation datasets, Citeseer, Cora-ML, and PubMed, and a co-authorship one, MS-Academic. Then we compare the performances of a subset of selected algorithms on Amazon Computer and Amazon Photo, that are segments of the Amazon co-purchase graph introduced in \cite{mcauley2015data}. All the datasets have a feature vector with a bag-of-words representation associated with the nodes. Other relevant characteristics are summarized in Table \ref{tab:datasets}.}
These features are normalized with an $\ell_1$ norm and to conclude the preprocessing, which is the same for all the datasets, we select the largest connected component.

\begin{table}[t]
\centering
\caption{\color{bostonuniversityred}Dataset statistics.}
\begin{tabular}{l|c|c|c|c|c}
Dataset & Classes & Features & Nodes & Edges & Avg. Degree \\
\hline
Citeseer & 6 & 3703 & 2110 & 3668 & 6.95\\
Cora-ML  & 7 & 2879 & 2810 & 7981 & 11.36 \\
PubMed  & 3 & 500 & 19717 & 44324  & 8.99 \\
MS-Academic & 15 & 6805 & 18333 & 81894 & 17.86 \\
A. Computers & 10 & 767 & 13381 & 245778 & 73.47  \\
A. Photos & 8 & 745 & 7487 & 119043 & 63.59
\end{tabular}
\label{tab:datasets}
\end{table}

To be in line with the evaluation of \cite{klicpera2018predict} we use the same number of layers ($2$) and hidden units ($64$), dropout rate ($0.5$) on both layers and the adjacency matrix, resampled at each propagation, and Adam optimizer \cite{kingma2014adam} with learning rate $0.01$. We choose instead the following hyperparameters for all the datasets: $\ell_2$ regularization parameter $0.008$ on the weights of the first layer, maximum steps of propagation $T=10$. {\color{bostonuniversityred} For the evaluation on Amazon's dataset we removed the $\ell_2$ regularization keeping the same learning rate for all the algorithms involved.} We adapted the propagation penalty $\alpha$, controlling the distribution of the propagation steps, to each dataset. 
The code to test our proposed AP-GCN and replicate our experiments is available on the web.\footnote{https://github.com/spindro/AP-GCN}  {\color{bostonuniversityred} The evaluation in \cite{klicpera2018predict} together with their proposed methods PPNP and APPNP included: GCN \cite{kipf2017semi}, both optimized and as originally proposed (V.GCN), network of GCNs (N-GCN) \cite{abu2018ncgn}, graph attention networks (GAT) \cite{velivckovic2017graph}, bootstrapped feature propagation (bt.FP) \cite{buchnik2018bs} and jumping knowledge networks with concatenation (JK) \cite{xu2018representation}}. We included in our comparison ARMA \cite{bianchi2019graph}, with a configuration compatible with the experimental setup.


\subsection{Results and comparisons}

\begin{figure}
    \centering
     \includegraphics[width=0.75\columnwidth]{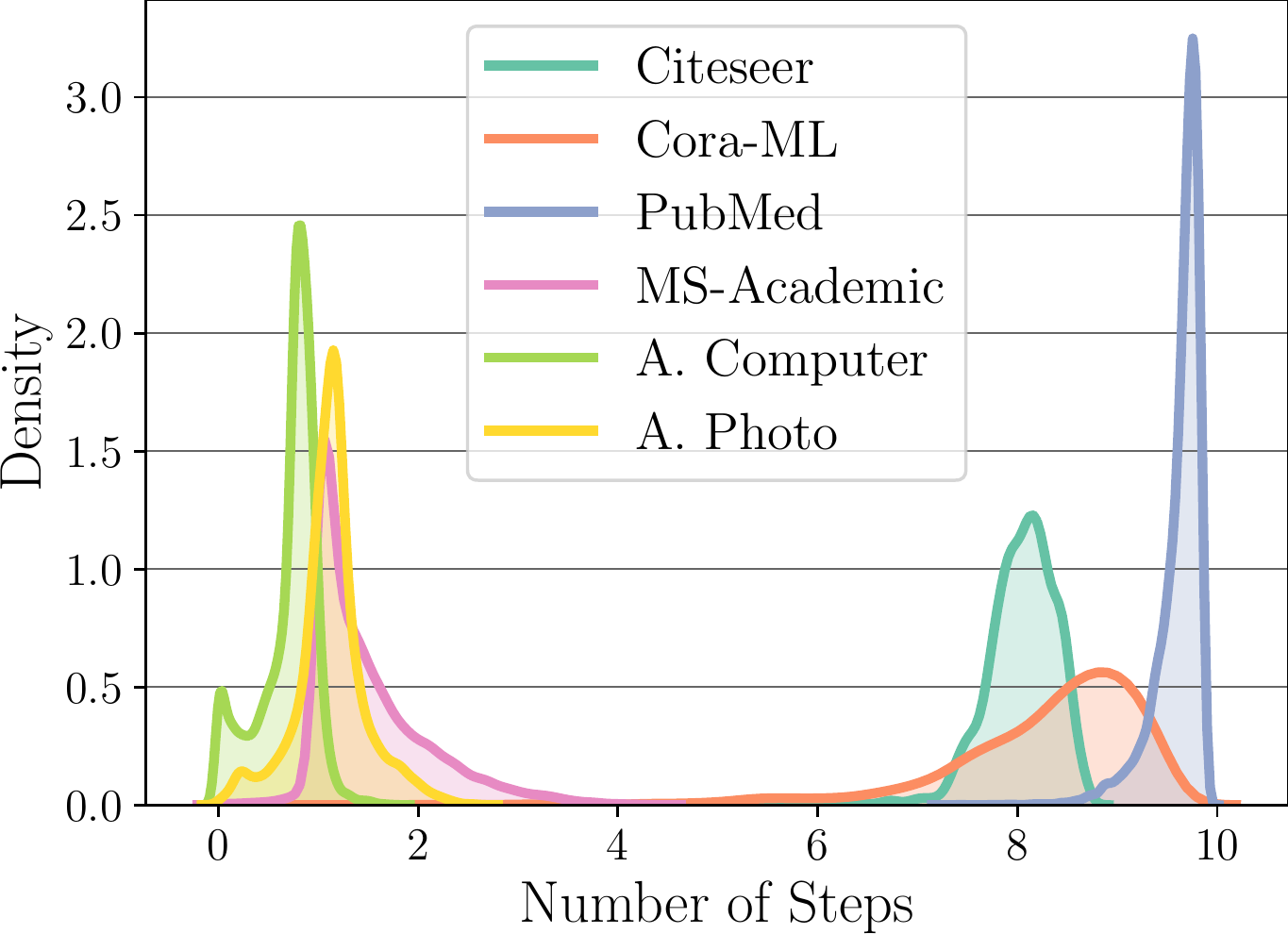}
    \caption{\color{bostonuniversityred} Average density distribution of the maximum number of propagations $K$ selected by AP-GCN in the evaluation associated to Table \ref{tab:node_class}.}
    \label{fig:density}
\end{figure}

In Tables \ref{tab:node_class} we report the average accuracy when using a training set of $20$ nodes per class, with uncertainties showing the $95\%$ confidence level calculated by bootstrapping. {\color{bostonuniversityred} In Table \ref{tab:node_class2} we use ($*$) and ($**$) to indicate statistical significance for a cutoff value of 0.05 and 0.01 respectively, when comparing the result to the second-best result using an aligned Friedman-rank test.}
{\color{bostonuniversityred} In Fig. \ref{fig:density} we show the distribution of the steps selected by AP-GCN. Fine-tuned values of $\alpha$ for each dataset are provided in Table \ref{tab:summary}. For Amazon's datasets, setting the APPNP restart probability to $0.2$ led to the best results.}

{\color{bostonuniversityred} AP-GCN outperforms its competitors over the citation graphs, meanwhile on the co-autorship graph APPNP remains the state-of-the-art. On the two Amazon datasets, which have very different characteristics, the improvements of AP-GCN are even more pronounced, and ARMA represents the second-best alternative.}
Furthermore, AP-GCN shows a low variance, which ensures robustness to the choice of the splits and random initializations.

\begin{table*}[t]
\centering
\caption{\color{bostonuniversityred}Average accuracy with uncertainties showing the 95\% confidence level calculated by boot-strapping.}
\resizebox{1.25\columnwidth}{!}{%
\begin{tabular}{l|c|c|c|c}
Model & Citeseer & Cora-ML &  PubMed & MS-Academic\\
\hline
V. GCN  & {73.51 $\pm$ 0.48} & {82.30 $\pm$ 0.34} & {77.65 $\pm$ 0.40} & {91.65 $\pm$ 0.09}  \\
GCN & {75.40 $\pm$ 0.30} & {83.41 $\pm$ 0.34} & {78.68 $\pm$ 0.38} & {92.10 $\pm$ 0.08} \\
N-GCN & {74.25 $\pm$ 0.40} & {82.25 $\pm$ 0.30} & {77.43 $\pm$ 0.42} & {92.86 $\pm$ 0.11} \\
GAT & {75.39 $\pm$ 0.47} & {84.37 $\pm$ 0.24} & {77.46 $\pm$ 0.44} & {91.22 $\pm$ 0.11} \\
JK & {73.03 $\pm$ 0.47} & {82.69 $\pm$ 0.35} & {77.88 $\pm$ 0.38} & {91.71 $\pm$ 0.07}  \\
Bt.FP & {73.55$\pm$ 0.57 } & {80.84 $\pm$ 0.97 } & {72.94 $\pm$ 1.00} & {91.61  $\pm$ 0.24} \\
PPNP & {75.83 $\pm$ 0.27} & {85.29 $\pm$ 0.25} & {-} & {-} \\
APPNP & {75.73 $\pm$ 0.30} & {85.09 $\pm$ 0.25} & {79.73 $\pm$ 0.31} & \textbf{93.27** $\pm$ 0.08} \\
ARMA   & {73.56 $\pm$ 0.36} & {82.58 $\pm$ 0.28} & 76.31 $\pm$ 0.41 & 92.41 $\pm$ 0.07  \\
AP-GCN & \textbf{76.12** $\pm$ 0.24} & \textbf{85.71** $\pm$ 0.22} & \textbf{79.80* $\pm$ 0.34} & {92.62 $\pm$ 0.07} 
\end{tabular}%
}
\label{tab:node_class}
\end{table*}

\begin{table}[t]
\centering
\caption{\color{bostonuniversityred}Average accuracy with uncertainties showing the 95\% confidence level calculated by boot-strapping.}
\begin{tabular}{l|c|c}
Model & A.Computer  & A.Photo \\
\hline
GCN & 78.62 $\pm$ 0.30 & 84.20 $\pm$  0.41 \\
GAT & 76.08 $\pm$ 0.47 & 88.21 $\pm$ 0.65 \\
APPNP & 80.17 $\pm$ 0.31  & 89.30 $\pm$ 0.24 \\
ARMA & 80.75 $\pm$ 0.37 & 89.48 $\pm$  0.33 \\
AP-GCN & \textbf{85.18** $\pm$ 0.23} & \textbf{92.05** $\pm$ 0.22}
\end{tabular}%
\label{tab:node_class2}
\end{table}

In Table \ref{tab:times} we report the average training time per epoch of our implementation of a subset of algorithms of Table \ref{tab:node_class} using the  framework introduced in \cite{fey2019pygeom}. Due to the higher number of propagation steps, and the presence of an additional (small) layer, AP-GCN {\color{bostonuniversityred} is among the slowest methods for smaller datasets. However, it scales better to bigger datasets with respect to GAT \cite{velivckovic2017graph}}. 

\begin{table}[t]
\centering
\caption{\color{bostonuniversityred} Selected $\alpha$ for each dataset, and corresponding average number of propagation steps. In the last column, we show the drop in accuracy across the range used for fine-tuning (see the text).}
\begin{tabular}{l|c|c|c}
Dataset & Best $\alpha$ & Avg. K (Best $\alpha$)& $\Delta$ Acc. $ (\alpha)$\\
\hline
Citeseer      & 0.001 & 8.85 $\pm$  0.31& 1.51 \\
Cora-ML       & 0.005 & 9.31 $\pm$  0.35& 4.65 \\
PubMed      & 0.001 & 9.62 $\pm$  0.17 & 2.44\\
MS-Academic   & 0.05 & 2.51 $\pm$  0.08& 0.51 \\
A.Computer   & 0.05 & 1.71 $\pm$ 0.06& 0.19 \\
A.Photo       & 0.05 & 2.13 $\pm$  0.05& 0.13
\end{tabular}
\label{tab:summary}
\end{table}

\begin{table}[t]
\centering
\caption{\color{bostonuniversityred} Average training time per epoch (milliseconds).}
\begin{tabular}{l|c|c|c|c|c}
Dataset & AP-GCN & ARMA & APPNP & GCN & GAT \\
\hline
Citeseer    & 32.4 & 25.2 & 19.6 & 8.6 & 11.1 \\
Cora-ML     & 36.2 & 27.6 & 22.1 & 7.9 & 13.4 \\
PubMed      & 42.0 & 51.1 & 23.3 & 16.1 & 45.4 \\
MS-Academic & 100.3 & 121.2 & 86.1 & 56.0 & 110.5 \\
A.Computer  & 76.7  & 80.0 & 76.7 & 50.2  & 222.6 \\
A.Photo     & 50.0 & 34.7 & 38.3  & 25.9 & 111.8 
\end{tabular}
\label{tab:times}
\end{table}

\subsection{Sensitivity to hyper-parameters}

\begin{figure}
    \centering
    \subfigure[Avg. Density Cora-ML]{\includegraphics[width=0.49\columnwidth]{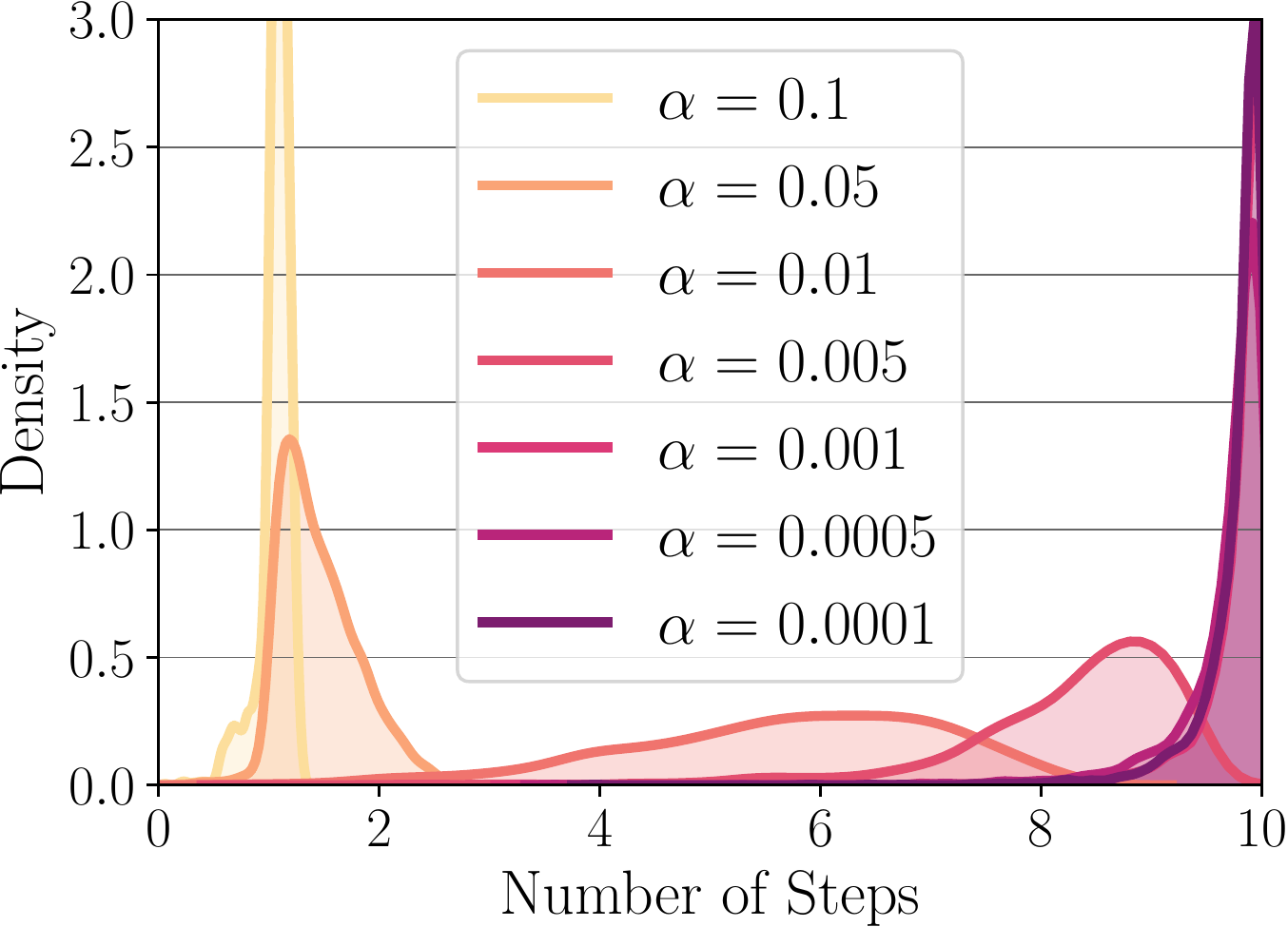}}
    \subfigure[Avg. Density A. Computer]{\includegraphics[width=0.49\columnwidth]{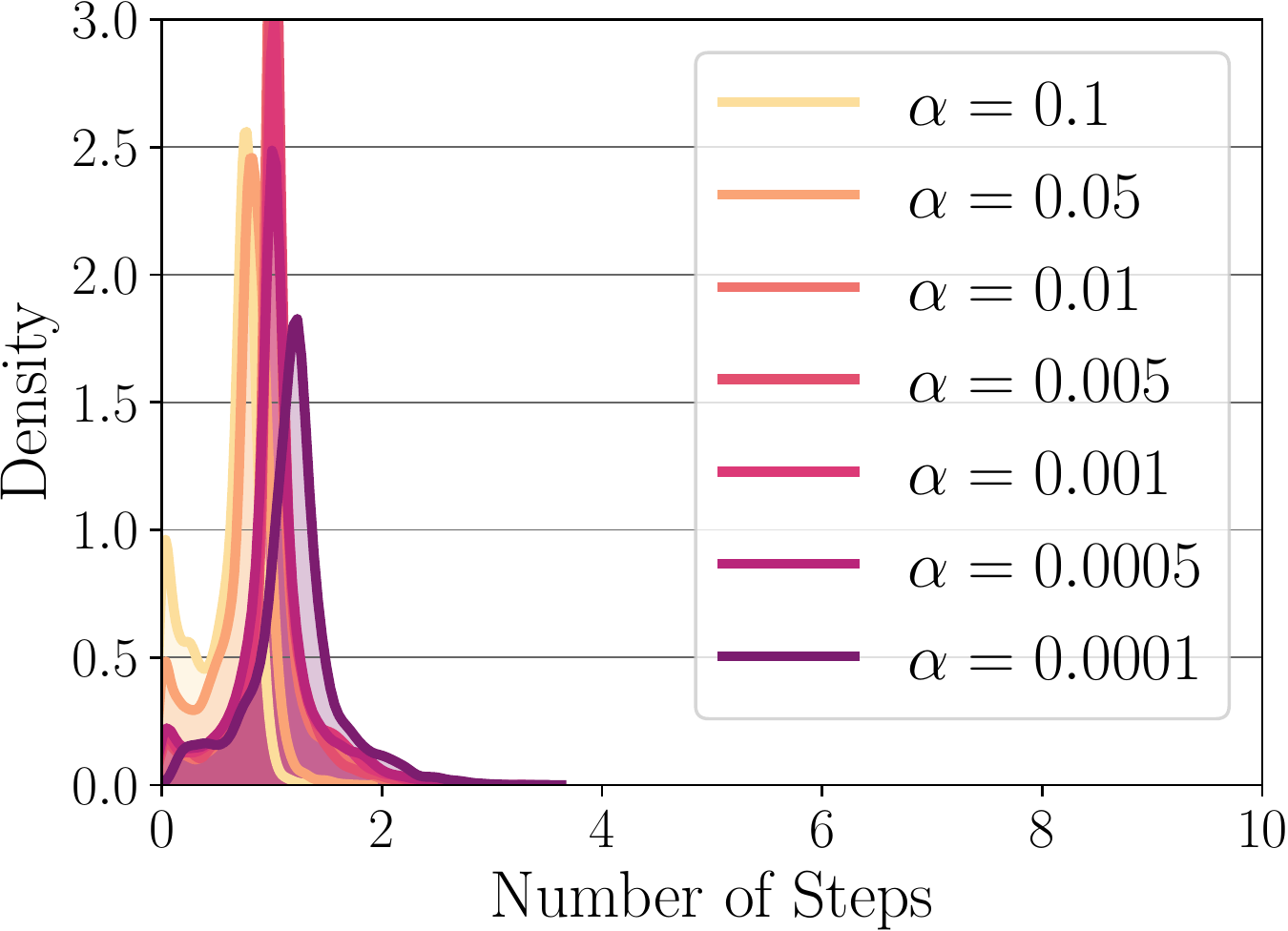}}
    \subfigure[Accuracy Cora-ML]{\includegraphics[width=0.49\columnwidth]{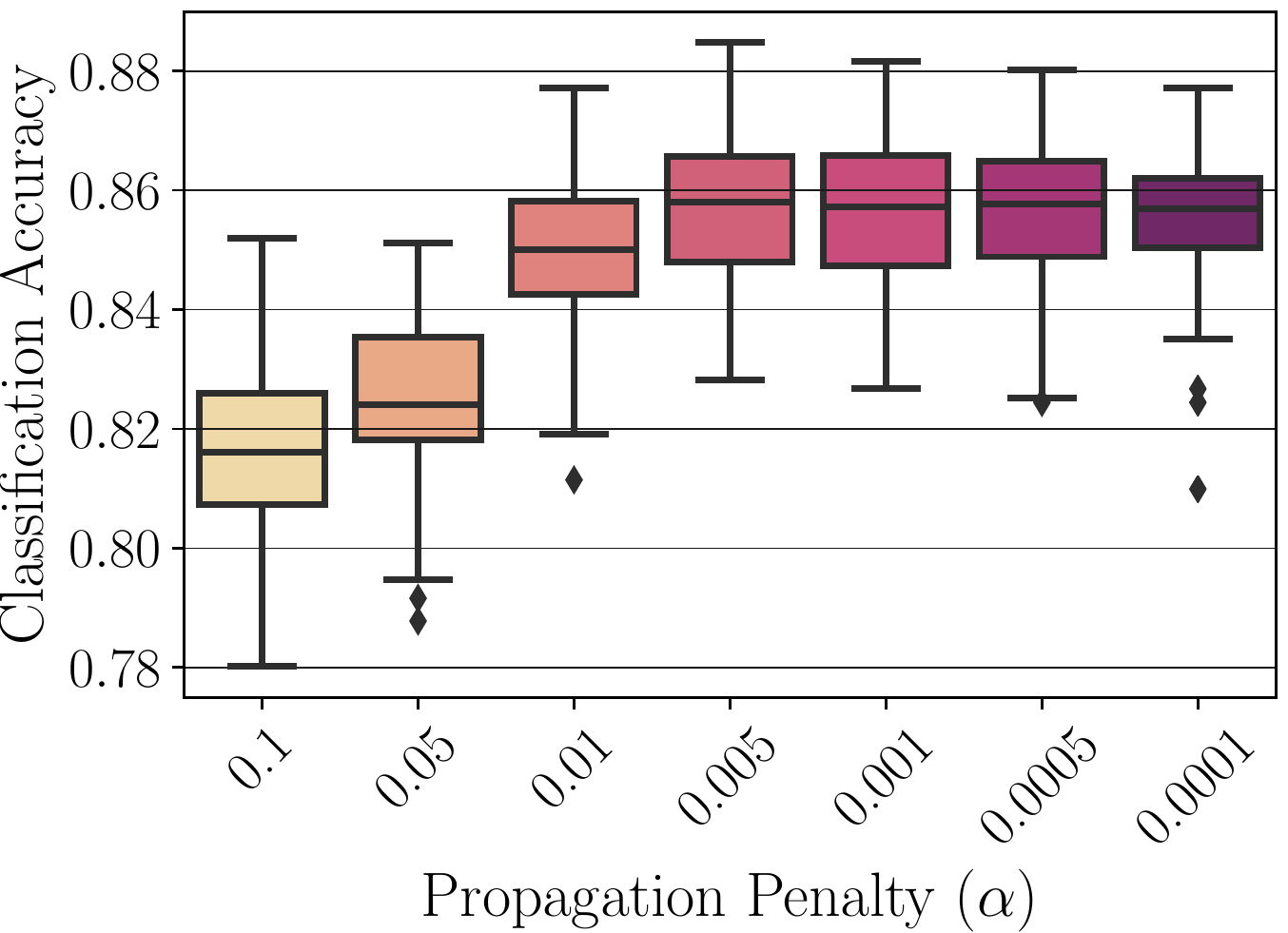}}
    \subfigure[Accuracy A. Computer]{\includegraphics[width=0.49\columnwidth]{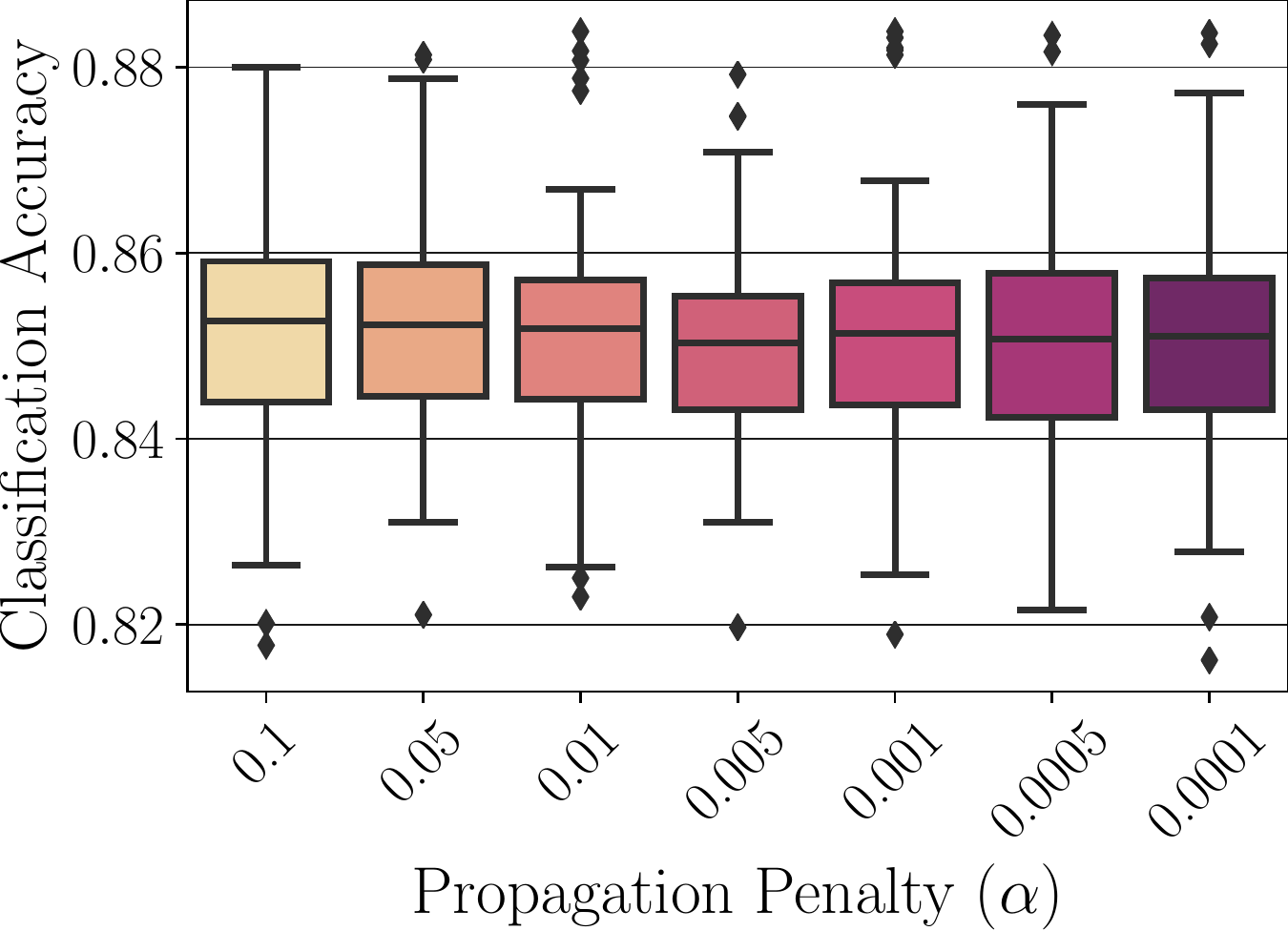}}
    \caption{\color{bostonuniversityred} (a)(b) Average density distribution of the maximum number of propagations $K$ and (c)(d) accuracy of AP-GCN on Cora-ML and Amazon Computer varying the propagation penalty $\alpha$ in the range $[0.1,0.0001]$.}
    \label{fig:alpha}
\end{figure}

Here we would like to inspect the sensitivity of AP-GCN to the propagation penalty $\alpha$. In Figure \ref{fig:alpha} {\color{bostonuniversityred} we show the variation in the average density distribution of the selected propagation steps and the corresponding accuracy. We selected two graphs with different characteristics that reflected the behaviour encountered in the other datasets. In any case, decreasing the value of the propagation penalty has the effect of augmenting the receptive field of AP-GCN. This is particularly useful in the case of nodes that are far away from labeled samples. A receptive field too big could lead to an over  smoothing problem and a consequent drop in performance.
The first dataset is Cora-ML (Figure \ref{fig:alpha} (a,c)), a relatively small dataset  with average degree once pre-processed of $11.36$. The variation of the propagation penalty in the range $[0.1,0.0001]$ lead to a selection of different optimal number of steps in the entire range $(0,10)$.} For higher values like $\alpha = 0.05$, AP-GCN performs mostly less than two propagation steps and the performances are comparable to the GCN reported in Table \ref{tab:node_class}. The best value for AP-GCN is found for $\alpha = 0.005$.
{\color{bostonuniversityred} The second dataset is Amazon Computer (Figure \ref{fig:alpha} (b,d)), a larger graph with an average degree of $73.47$. The variation of $\alpha$ in this case has a limited effect over the range of selected propagation steps. In fact, all the average densities lie in the range $(0,4)$ and the variation of the accuracy is less pronounced. This is most likely due to large degree, that translates into a greater amount of information transmitted at every propagation step. This could lead to an over-smoothing effect but AP-GCN, robustly with respect to the choice of $\alpha$, adapts itself to the characteristics of the graph to avoid this issue.}

\begin{figure}
    \centering
    \subfigure[Accuracy]{\includegraphics[width=0.49\columnwidth]{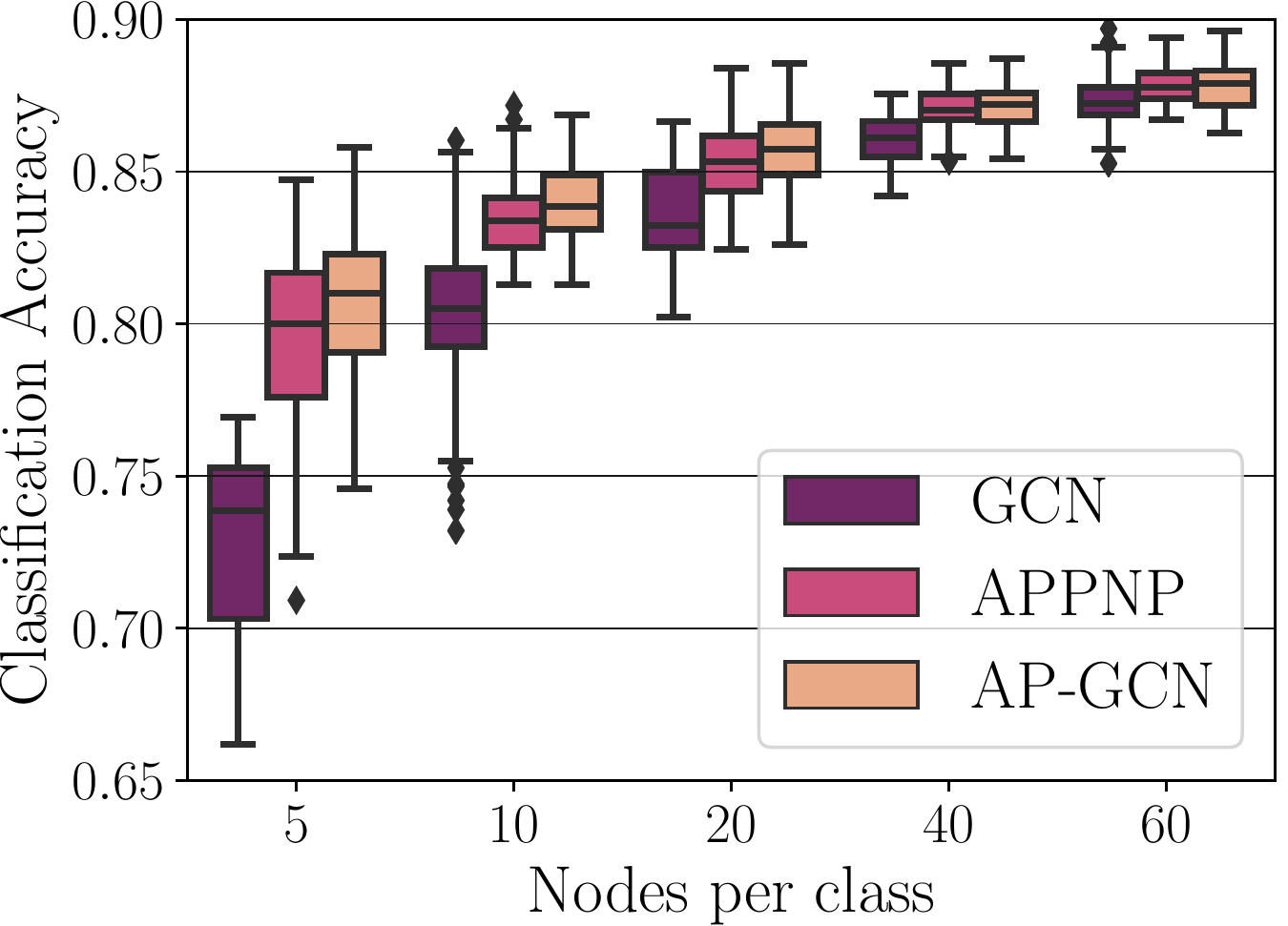}}
    \subfigure[Avg. Density]{\includegraphics[width=0.49\columnwidth]{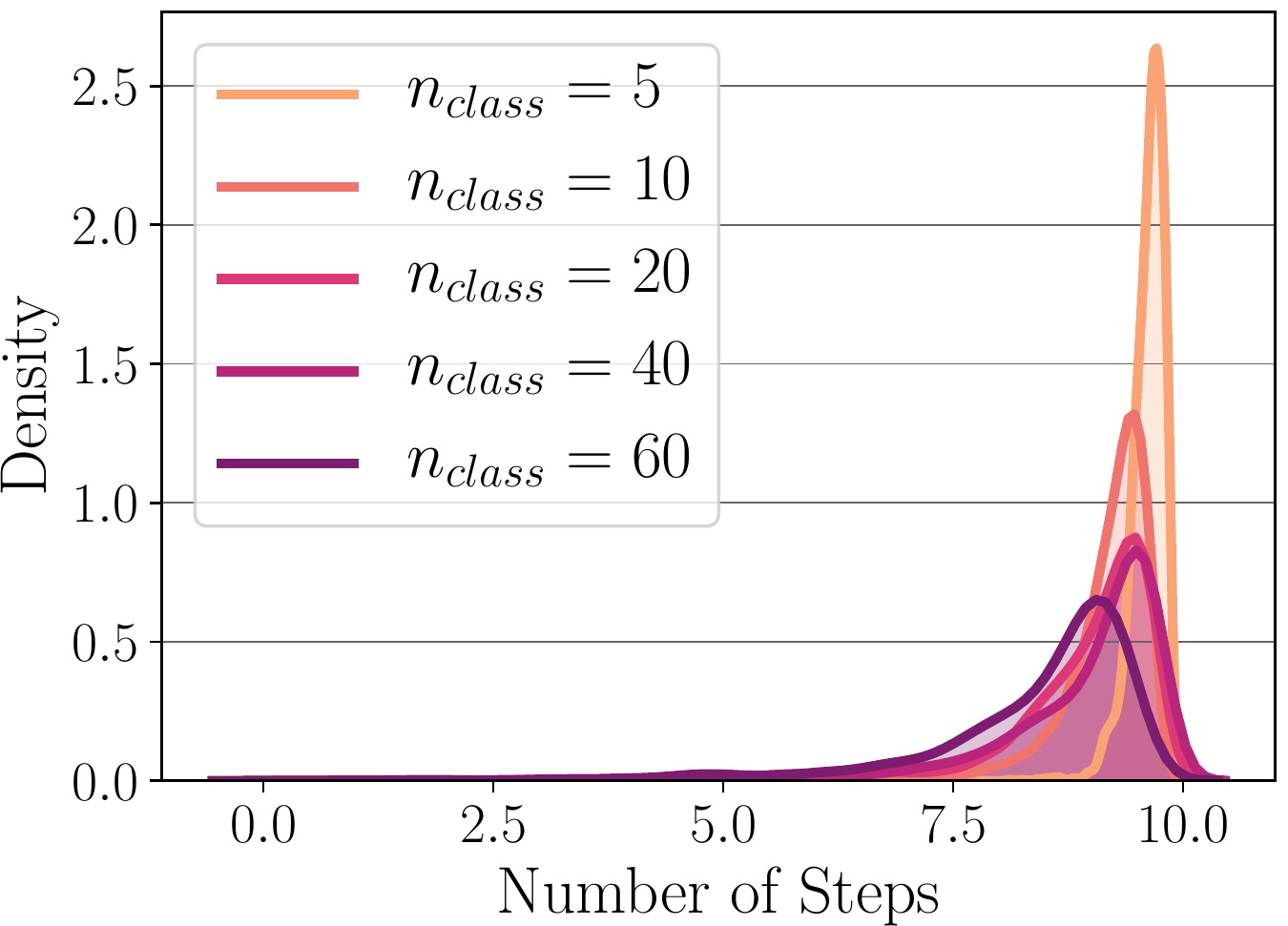}}
    \caption{(a) Accuracy of GCN, APPNP and AP-GCN  for different numbers of labeled nodes per class on Cora-ML. (b) AP-GCN relative average density distribution of the maximum number of propagations $K$.}
    \label{fig:nclass}
\end{figure}

Finally, we want to analyze the performances of GCN, APPNP, and AP-GCN on Cora-ML for different dimensions of the training set. This is a crucial aspect since labelling is one of the most expensive processes in modern machine learning. Therefore a model capable of working with very few labelled samples has a great advantage over those that do not. Fig. \ref{fig:nclass}(a) shows, as noticed in \cite{klicpera2018predict}, that the higher range of APPNP and AP-GCN permits to have a great increment in performance when the label information is very sparse. The improvement of AP-GCN over APPNP, even if present for every size of the training set, behaves similarly. This suggests that a loosely labelled dataset highlights the effectiveness of a propagation protocol. In Figure \ref{fig:nclass}(b) we show the variation of the average density distribution of the maximum number of propagation steps selected by AP-GCN under the different training sizes. The behaviour of AP-GCN is in line with the previous observation. The sparsest the labels, the more propagation steps performed by AP-GCN, trying to spread this information. Contrary, when the number of labelled samples increases more and more, nodes in the graph select as maximum propagation $ K < T$, preventing the issue of over-smoothing.

\section{Conclusion}
\label{sec:conclusion}
In this paper we introduced the adaptive propagation graph convolutional network (AP-GCN), a variation of GCN wherein each node selects automatically the number of propagation steps performed across the graph. We showed experimentally that the method performs favourably or better than the state-of-the-art, that it is robust to the training set size and,  {\color{bostonuniversityred}in most cases, it can adapt its behaviour to the dataset more or less robustly depending on the hyper-parameter's choice.} Future work will consider extending the ideas presented here to different types of GNNs and to tasks going beyond node classification. {\color{bostonuniversityred}Our update is similar to the PageRank and ARMA models proposed in \cite{klicpera2018predict,bianchi2019graph}, which are known to approximate a rational filter on the graph \cite{chen2020bridging}. Future work will also explore in-depth the spectral properties of our model.}

\bibliographystyle{IEEEtran}
\bibliography{biblio}

\end{document}